\documentclass[twoside]{article}

\usepackage[accepted]{aistats2022}
\usepackage{amsmath}
\usepackage{amssymb}
\usepackage{amsthm}
\usepackage{mathtools}
\usepackage{bm}
\usepackage[round]{natbib}
\usepackage[inline]{enumitem}
\usepackage{tikz}
\usepackage{booktabs}
\usepackage{multirow}
\usepackage{graphicx}
\usepackage{subcaption}
\usepackage{mwe}

\usepackage{array}
\newcommand{\PreserveBackslash}[1]{\let\temp=\\#1\let\\=\temp}
\newcolumntype{C}[1]{>{\PreserveBackslash\centering}p{#1}}
\newcolumntype{R}[1]{>{\PreserveBackslash\raggedleft}p{#1}}
\newcolumntype{L}[1]{>{\PreserveBackslash\raggedright}p{#1}}

\usepackage{color}
\usepackage{colortbl}
\definecolor{deepblue}{rgb}{0,0,0.5}
\definecolor{deepred}{rgb}{0.6,0,0}
\definecolor{deepgreen}{rgb}{0,0.5,0}
\definecolor{gray}{rgb}{0.7,0.7,0.7}

\usepackage{hyperref}
\hypersetup{
  colorlinks   = true, 
  urlcolor     = black, 
  linkcolor    = blue, 
  citecolor    = blue  
}

\newtheorem{assumption}{Assumption}

\newtheorem{lemma}{Lemma}
\newtheorem{theorem}{Theorem}
\newtheorem{corollary}{Corollary}

\newcommand{\R}{\mathbb R}

\DeclareMathOperator{\E}{\mathbb E}

\DeclareMathOperator*{\argmin}{arg\,min}

\newcommand{\parent}[1]{\texttt{parent}({#1})}

\renewcommand{\star}[1]{{#1}^{*}}
\newcommand{\bad}[1]{{#1}^{\textit{bad}}}
\newcommand{\trans}[1]{{#1}^{T}}

\newcommand{\vv}{\mathbf v}
\newcommand{\uu}{\mathbf u}
\newcommand{\w}{\mathbf w}
\newcommand{\x}{\mathbf x}

\newcommand{\ltwo}[1]{{\lVert {#1} \rVert}_2}

\newcommand{\lF}[1]{{\lVert {#1} \rVert}_F}

\newcommand{\depth}[1]{\texttt{depth}({#1})}

\newcommand{\ignore}[1]{}

\begin{document}

%

%

\twocolumn[

\aistatstitle{The Tree Loss: Improving Generalization with Many Classes}

\aistatsauthor{ Yujie Wang \And Mike Izbicki  }

\aistatsaddress{ Claremont Graduate University \And  Claremont McKenna College } ]

\begin{abstract}

    Multi-class classification problems often have many semantically similar classes.
    For example, 90 of ImageNet's 1000 classes are for different breeds of dog.
    We should expect that these semantically similar classes will have similar parameter vectors,
    but the standard cross entropy loss does not enforce this constraint.

    We introduce the tree loss as a drop-in replacement for the cross entropy loss.
    The tree loss re-parameterizes the parameter matrix in order to guarantee that semantically similar classes will have similar parameter vectors.
    Using simple properties of stochastic gradient descent,
    we show that the tree loss's generalization error is asymptotically better than the cross entropy loss's.
    We then validate these theoretical results on synthetic data, image data (CIFAR100, ImageNet), and text data (Twitter).


\end{abstract}

\section{Introduction}


The cross entropy loss is the most widely used loss function for multi-class classification problems,
and stochastic gradient descent (SGD) is the most common algorithm for optimizing this loss.
Standard results show that the generalization error decays at a rate of $O(\sqrt{k/n})$,
where $k$ is the number of class labels,
and $n$ is the number of data points.
These results (which we review later) make no assumptions about the underlying distribution of classes.

In this paper,
we assume that the class labels have an underlying metric structure,
and we introduce the tree loss for exploiting this structure.
The doubling dimension $c$ of a metric space is a common measure of the complexity of the metric,
and we show that SGD applied to the tree loss will converge at a rate of $O(\sqrt{\log k/n})$ when $c\le 1$ or $O(\sqrt{k^{1-1/c}/n})$ when $c\ge 1$.
These improvements are obviously most dramatic for small $c$,
and the tree loss outperforms the cross entropy loss best in this regime.
The tree loss is the first multi-class loss function with provably better generalization error than the cross entropy loss in any regime.

Our paper is organized as follows.
We begin in Section \ref{sec:related} by discussing limitations of related loss functions and how the tree loss addresses those limitations.
Section \ref{sec:problem} then formally defines the problem setting, and Section \ref{sec:tree} formally defines the tree loss.
We emphasize that the tree loss is simply a reparameterization of the standard cross entropy loss,
and so it is easy to implement in common machine learning libraries.
Section \ref{sec:theory} reviews standard results on the convergence of stochastic gradient descent,
and uses those results to prove the convergence bounds for the tree loss.
Finally, Section \ref{sec:experiment} conducts experiments on synthetic, real world image (CIFAR100, ImageNet) and text (Twitter) data.
We show that in practice, the tree loss essentially always outperforms other multi-class loss functions.

\begin{figure*}
\resizebox{\columnwidth}{!}{
\begin{tikzpicture}
    [ level distance=1.5cm
    , level 1/.style={sibling distance=3cm}
    , level 2/.style={sibling distance=1.5cm}
    ]
\node[draw, rounded corners=0.1in, inner sep=0.1in] at (-1.5in,0){\textbf{$U$-Tree}};
\node {\texttt{truck}}
    child {node {\texttt{boxer}}
      child {node {\texttt{tiger}}}
      child {node {\texttt{skunk}}}
      child {node {\texttt{boxer}}
        child {edge from parent[draw=none]} 
        child {node {\texttt{bulldog}}}
        child {node {\texttt{boxer}}}
        child {node {\texttt{husky}}}
        child {node {\texttt{sheepdog}}}
      }
      child {node {\texttt{bear}}}
      child {edge from parent[draw=none]} 
      }
    child {node {\texttt{truck}}
      child {edge from parent[draw=none]} 
      child {node {\texttt{truck}}}
      child {node {\texttt{bus}}}
    }
    child {node {\texttt{toaster}}}
    ;

\end{tikzpicture}
}
~~~
\resizebox{\columnwidth}{!}{
\begin{tikzpicture}
    [ level distance=1.5cm
    , level 1/.style={sibling distance=3cm}
    , level 2/.style={sibling distance=1.5cm}
    ]
\node[draw, rounded corners=0.1in, inner sep=0.1in] at (-1.5in,0){\textbf{$V$-Tree}};
\node {\textit{pseudo1}}
    child {node {\textit{pseudo2}}
      child {node {\texttt{tiger}}}
      child {node {\texttt{skunk}}}
      child {node {\textit{pseudo3}}
        child {edge from parent[draw=none]} 
        child {node {\texttt{bulldog}}}
        child {node {\texttt{boxer}}}
        child {node {\texttt{husky}}}
        child {node {\texttt{sheepdog}}}
      }
      child {node {\texttt{bear}}}
      child {edge from parent[draw=none]} 
      }
    child {node {\textit{pseudo4}}
      child {edge from parent[draw=none]} 
      child {node {\texttt{truck}}}
      child {node {\texttt{bus}}}
    }
    child {node {\texttt{toaster}}}
    ;

\end{tikzpicture}
}
    \caption{
        Example label tree structures for a subset of 10 ImageNet classes.
        The $U$-tree has class labels that repeat at multiple levels,
        and the $V$-tree introduces ``pseudoclasses''.
        The pseudoclass \textit{pseudo3} represents the class of ``dogs'',
        and the pseudoclass \textit{pseudo2} represents the class of ``animals''.
    }
    \label{fig:labeltree}
\end{figure*}
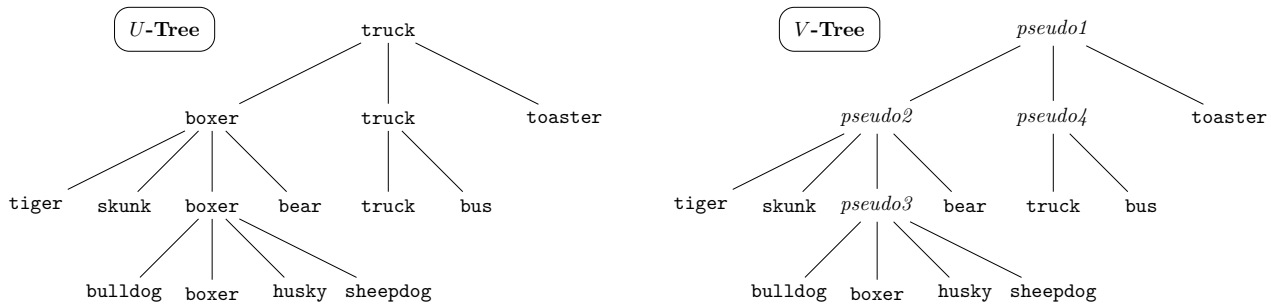


\section{Related Work}
\label{sec:related}

Previous work on multi-class loss functions has focused on improving either the statistical or computational performance.
Statistical work includes loss functions designed for hierarchichally structured labels \citep{cesa2006incremental,wu2017hierarchical,bertinetto2020making},
loss functions for improving top-$k$ accuracy \citep{lapin2016loss},
or focusing on noisy labels \citep{sukhbaatar2014training,zhang2018generalized}.
The work most similar to ours is the SimLoss \citep{Kobs2020SimLossCS},
which also makes a metric assumption about the class label structure.
Our work improves on all of this statistical work by being the first to provide convergence bounds that improve on the bounds of the cross entropy loss.

Other work focuses on improving the speed of multiclass classification.
The hierarchical softmax \citep{morin2005hierarchical} is a particularly well studied modification of the cross entropy loss with many variations \citep[e.g.][]{Peng2017IncrementallyLT,Jiang2017ExplorationOT,Yang2017OptimizeHS,Mohammed2018EffectivenessOH}.
It is easily confused with our tree loss because both loss functions involve a tree structure.
The difference, however, is that the hierarchical softmax focuses on improving runtime performance,
and most variants actually sacrifice statistical performance to do so.
The tree loss, in contrast, maintains the runtime performance of the standard cross entropy loss but improves the statistical performance.


\section{Problem Setting}
\label{sec:problem}

We consider the multi-class classification setting with $k$ classes and $d$ input feature dimensions.
The cross entropy loss is the standard loss function for this setting.
It is defined to be
\begin{equation}
    \label{eq:xentropy}
    \ell(W;(\x,y)) = - \log \frac {\exp(-\trans\w_y \x)}{\sum_{j=1}^k \exp(-\trans \w_j \x)}
\end{equation}
where for each class $i\in[k]$,
$\w_i : \R^d$ is the parameter vector associated with class $i$;
the variable $W : \R^{k \times d} = (\w_1; \w_2; ...; \w_k)$ is the full parameter matrix;
$\x : \R^d$ is the input feature vector;
and $y \in [k]$ is the input class label.

The cross entropy loss has no constraints on the weight matrix that cause similar classes to have similar parameter vectors.
The tree loss adds these constraints,
resulting in faster convergence and better generalization.


\section{The Tree Loss}
\label{sec:tree}

The main idea of the tree loss is that similar classes should be forced to ``share'' some of their parameters.
The tree loss refactors the cross entropy loss's weight matrix in order to enforce this sharing.

We propose two variants of the tree loss which enforce this sharing in different ways.
We begin by introducing the $U$-tree loss,
which is simpler to explain and analyze.
Then, we introduce the $V$-tree loss,
which improves on the $U$-tree loss and is the loss we suggest using in practice.
Whenever we refer to the tree loss without a qualifier,
we mean the $V$-tree loss.



\subsection{The $U$-Tree Loss}

Explaining our new $U$-tree parameterization requires introducing some notation.

We define a $U$-tree over the class labels to be a tree where each leaf node is represented by a label,
and each internal node shares a label with a child node.
Figure \ref{fig:labeltree} (left) shows an example $U$-tree structure.
The definition of a $U$-tree is motivated by the cover tree data structure \citep{beygelzimer2006cover,izbicki2015faster},
which generates $U$-tree structures given a metric over the class labels.

For each class $i$, we let the sequence $P_i$ denote the path from the leaf node to the root with duplicates removed.
For example, using the $U$-tree in Figure \ref{fig:labeltree},
we have
\begin{equation*}
\begin{split}
    P_{\texttt{bear}} &= (\texttt{bear}, \texttt{boxer}, \texttt{truck}) \\
    P_{\texttt{sheepdog}} &= (\texttt{sheepdog}, \texttt{boxer}, \texttt{truck}) \\
    P_{\texttt{truck}} &= (\texttt{truck})
\end{split}
\end{equation*}
For each class $i$, we define $\parent{i}$ to be the first node in $P_i$ not equal to $i$.
In other words, $\parent{i}$ is the parent of the upper-most node for class $i$.

We are now ready to introduce the $U$-tree parameterization.
We associate for each class $i$ a new parameter vector $\uu_i$ defined to be
\begin{equation}
    \uu_i =
    \begin{cases}
        \w_i                    & \text{if $i$ is the root} \\
        \w_{\parent{i}} - \w_i  & \text{otherwise}
    \end{cases}
    ,
\end{equation}
and we define the parameter matrix $U : \mathbb R^{k\times d}$ to be $(\uu_1;\uu_2;...;\uu_k)$.
We can recursively rewrite the original parameter vectors in terms of this new parameterization as
\begin{equation}
    \w_i = \sum_{j\in P_i} \uu_i
    .
    \label{eq:uu}
\end{equation}
The $U$-tree loss is then defined by substituting \eqref{eq:uu} into \eqref{eq:xentropy} to get
\begin{equation*}
\ell(U;(\x,y)) = - \log \frac {\exp(-\sum_{j\in P_y}\trans\uu_j \x)}{\sum_{i=1}^k \exp(- \sum_{j\in P_i}\trans\uu_j \x)}
\end{equation*}
We use the same $\ell$ notation for the standard cross entropy loss function and the $U$-Tree loss because the function is exactly the same;
the only difference is how we represent the parameters during the training procedure.

\subsection{The $V$-Tree Loss}

The $V$-tree is constructed from the $U$-tree by replacing all non-leaf nodes with ``pseudoclasses''.
Let $k'$ denote the total number of classes plus the total number of pseudoclasses.

We let $P_i$ be defined as the path from leaf node $i$ to the root as before,
but note that there are now no duplicates to remove and so the paths are longer.
For example,
\begin{equation*}
\begin{split}
    P_{\texttt{bear}} &= (\texttt{bear}, \textit{pseudo2}, \textit{pseudo1}) \\
    P_{\texttt{sheepdog}} &= (\texttt{sheepdog}, \textit{pseudo3}, \textit{pseudo2}, \textit{pseudo1} ) \\
    P_{\texttt{truck}} &= (\texttt{truck}, \textit{pseudo4}, \textit{pseudo1} )
\end{split}
\end{equation*}
The $\vv_i$ and $V$ parameters are now defined analogously to the $\uu_i$ and $U$ parameters, but over the $V$-tree structure instead of the $U$-tree structure.
An important distinction between the $V$ and $U$ parameter matrices is that they will have different shapes.
The $U$ matrix has shape $k \times d$, which is the same as the $W$ matrix,
but the $V$ matrix has shape $k' \times d$,
which is potentially a factor of up to 2 times larger.

The $V$-tree loss is defined by using the $V$ matrix to parameterize the cross entropy loss, giving
\begin{equation}
    \ell(V;(\x,y)) = - \log \frac {\exp(-\sum_{k\in P_y}\trans\vv_k \x)}{\sum_{j=1}^k \exp(- \sum_{k\in P_j}\trans\vv_k \x)}
    .
\end{equation}
We continue to use the $\ell$ function to represent both the standard cross entropy loss and the $V$-tree loss function to emphasize that these are the same loss functions,
just with different parameterizations of the weight matrix.

\subsection{Intuition}

The intuition behind the $V$-tree reparameterization is that whenever we perform a gradient update on a data point with class $i$,
we will also be ``automatically'' updating the parameter vectors of similar classes.
To see this, note that when we have a \texttt{sheepdog} data point,
we will perform gradient updates on all $\vv_i$ in $P_\texttt{sheepdog}$; i.e. $\vv_{sheepdog}$, $\vv_{\textit{pseudo3}}$, $\vv_{\textit{pseudo2}}$, and $\vv_{\textit{pseudo1}}$.
This will cause the $\w_{\texttt{husky}}$ and $\w_{\texttt{bear}}$ parameters (among many others) to update because they also depend on the pseudoclass vectors $\vv_{\textit{pseudo3}}$ and $\vv{_\textit{pseudo2}}$.
Because $P_\texttt{husky}$ has a larger overlap with $P_\texttt{sheepdog}$ than $P_\texttt{bear}$,
the parameters of this more similar class will be updated more heavily.

This reparameterization is reminiscent of the way that fastText \citep{bojanowski2017enriching} reparameterizes the word2vec \citep{Mikolov2013EfficientEO} model to improve statistical efficiency.
Two notable differences, however, are that fastText is a domain-specific adaptation and provides no theoretical guarantees;
our tree loss works on any domain and provides theoretical guarantees.

\subsection{Implementation Notes}

Both the $U$-tree and $V$-tree losses are easy to implement in practice using the built-in cross entropy loss function in libraries like PyTorch \citep{NEURIPS2019_9015} or Tensorflow \citep{tensorflow2015-whitepaper}.
The only change needed is to represent the $W$ parameter matrix in your code as an appropriate sum over the $U$ or $V$ parameter matrices.
The magic of automatic differentiation will then take care of the rest and ensure that the underlying $U$ or $V$ matrix is optimized.
Our TreeLoss library\footnote{\url{https://github.com/cora1021/TreeLoss}} provides easy-to-use functions for generating these matrices,
and so modifying code to work with the tree loss is a 1 line code change.

In practice, the tree loss is slightly slower than the standard cross entropy loss due to the additional summation over the paths.
In our experiments, we observed an approximately 2x slowdown.
Modifications to the cross entropy loss that improve runtime (such as the hierarchical softmax or negative sampling) could also be used to improve the runtime of the tree loss,
but we do not explore this possibility in depth in this paper.


\section{Theoretical Results}
\label{sec:theory}

We use standard properties of stochastic gradient descent to bound the generalization error of the tree loss.
In this section we formally state our main results,
but we do not review the details about stochastic gradient descent or prove the results.
The Appendix reviews stochastic gradient descent in detail and uses the results to prove the theorems below.

To state our results, we first must introduce some learning theory notation.
Define the true loss of our learning problem as
\begin{equation}
    L_D(A) = \E_{(\x,y)\sim D} \ell(A; (\x, y))
\end{equation}
where $A \in \{W, U, V\}$ is a parameterization of the cross entropy loss.
We define $\bar A$ to be the result of running SGD on $n$ data points to optimize parameterization $A$,
and 
\begin{equation}
    \star A = \argmin_{A} L_D(A)
\end{equation}
to be the optimal parameterization.
Our goal is to bound the generalization error
\begin{equation}
    \E L_D(\bar A) - L_D(\star A).
\end{equation}
In order to bound this error, we make the following standard assumption about our data.
\begin{assumption}
    \label{ass:lip}
    For each feature vector $\x$ in the data set, 
        $\ltwo{\x} \le \rho$
        .
\end{assumption}
This assumption is equivalent to stating that $L_D$ (or $\ell$) is $\rho$-Lipschitz.

Now our key observation is that $\lF{\star A}$ bounds the generalization error,
as formalized in the following lemma.
\begin{lemma}
    \label{ref:cor:A}
    Under Assumption \ref{ass:lip},
    we have that for any parameterization $A$ of the cross entropy loss,
    \begin{equation}
        \E L_D(\bar A) - L_D(\star A) \le \frac{\lF{\star A}\rho}{\sqrt n}.
        \label{eq:Aconv}
    \end{equation}
\end{lemma}


We will next show how to use Lemma \ref{ref:cor:A} to recover the standard convergence rate of multi-class classification by bounding $\lF{\star W}$.
We make the following assumption.
\begin{assumption}
    \label{ass:B}
    For each class $i$, the optimal parameter vector $\star\w_i$ satisfies
        $\ltwo{\star\w_i} \le B$
        .
\end{assumption}
It follows that
\begin{equation}
    \lF{\star W}^2 = \sum_{i=1}^k \ltwo{\star\w_i} \le kB^2.
    \label{eq:starW}
\end{equation}
Substituting Eq \eqref{eq:starW} into \eqref{eq:Aconv} gives the following bound.
\begin{corollary}
\label{theorem:xentropy}
    Under assumptions \ref{ass:lip} and \ref{ass:B},
    then the generalization error of the standard cross entropy parameterization when trained with SGD is
\begin{equation}
    \E L_D(\bar W) - L_D(W^*)
    \le \frac {\sqrt kB\rho}{\sqrt n}
    .
\end{equation}
\end{corollary}

Next, we bound $\lF{\star U}$ and $\lF{\star V}$ in order to bound the generalization error of the $U$/$V$ parameterizations.
The analysis is divided into two parts.
First, we make no assumptions that the $U$-tree structure is good,
and we show that even in the worst case $\lF{\star U} = O(\lF{\star W})$.
This implies that using the tree loss cannot significantly hurt our performance.
\begin{lemma}
    \label{lemma:starU}
    Under assumption \ref{ass:B},
    the following bound holds for all $U$/$V$-tree structures:
    \begin{equation}
        \lF{\star V} \le \lF{\star U} \le 2\sqrt{k}B.
    \end{equation}
\end{lemma}
Now we consider the more important case of when we have a tree structure that meaningfully captures the similarity between classes.
This idea is captured in our final assumption.
\begin{assumption}
    \label{ass:metric}
    Let $\lambda \ge 1$, and $d$ be a distance metric over the labels such that for all labels $i$ and $j$,
\begin{equation}
    \frac 1 \lambda d(i,j)
    \le \ltwo{\star \w_i - \star \w_j}
    \le \lambda d(i, j).
\end{equation}
We denote by $c$ the doubling dimension of the metric $d$,
and we assume that the $U$-tree structure is built using a cover tree \citep{beygelzimer2006cover}.
\end{assumption}

The $\lambda$ parameter above measures the quality of our metric.
When $\lambda=1$, the metric $d$ is good and perfectly predicts the distance between parameter vectors;
when $\lambda$ is large, the metric $d$ is bad.

The Cover Tree was originally designed for speeding up nearest neighbor queries in arbitrary metric spaces.
The definition is rather technical, so we do not restate it here.
Instead, we mention only two important properties of the cover tree.
First, it can be constructed in time $O(k)$.
This is independent of the number of training examples $n$,
so building the $U$-tree structure is a cheap operation that has no meaningful impact on training performance.
Second, the cover tree has a hyperparameter which we call \texttt{base}.
This hyperparameter controls the fanout and depth of the tree because at any node $i$ at depth $\depth{i}$ in the tree,
the cover tree maintains the invariant that $d(i, \parent{i}) \le \texttt{base}^{-\depth{i}}$.
Increasing the $\texttt{base}$ hyperparameter results in shorter, fatter trees,
and decreasing it results in taller, narrower trees.
Our analysis follows the convention of setting $\texttt{base}=2$,
but we show in the experiments below that good performance is achieved for a wide range of base values.

The following Lemma uses Assumption \ref{ass:metric} and properties of the cover tree to bound the norm of $\lF{\star U}$.
It is our most difficult technical result.
\begin{lemma}
    \label{lemma:main}
    Under Assumptions \ref{ass:B} and \ref{ass:metric},
    when $c\le1$, we have that
    \begin{equation}
        \lF{\star U} \le \tfrac{1}{\sqrt2}\lambda B \sqrt{\log_2 k},
        \label{eq:c<=1}
    \end{equation}
    and when $c>1$, we have that
    \begin{equation}
        \lF{\star U} \le \sqrt{5}\lambda B \sqrt{k^{(1-1/c)}}.
        \label{eq:c>1}
    \end{equation}
\end{lemma}

We note that embedding techniques can be used to reduce the intrinsic dimension of the metric ($c$) at the expense of increasing the metric's distortion ($\lambda$),
but we make no attempt to optimize this tradeoff in this paper.

We now state our main result.
It is an immediate consequence Lemmas \ref{ref:cor:A}, \ref{lemma:starU} and \ref{lemma:main}.
\begin{corollary}
    \label{cor:main}
    Under assumptions \ref{ass:lip}, \ref{ass:B}, and \ref{ass:metric},
    when $c\le1$, the generalization error of the tree loss is bounded by
\begin{equation}
    \E L_D(\bar V) - L_D(V^*)
    \le \frac {\lambda B\rho \sqrt{\log_2 k}}{\sqrt 2n}
    .
\end{equation}
    When $c>1$, the generalization error of the tree loss is bounded by
\begin{equation}
    \E L_D(\bar V) - L_D(V^*)
    \le \frac {\lambda B\rho \sqrt{5 k^{(1-1/c)}}}{\sqrt n}
    .
\end{equation}
\end{corollary}
These convergence rates are asymptotically better than the convergence rates for the standard parameterization of the cross entropy loss.


\begin{figure*}
    \centering 

\includegraphics[width=\columnwidth,height=1.45in]{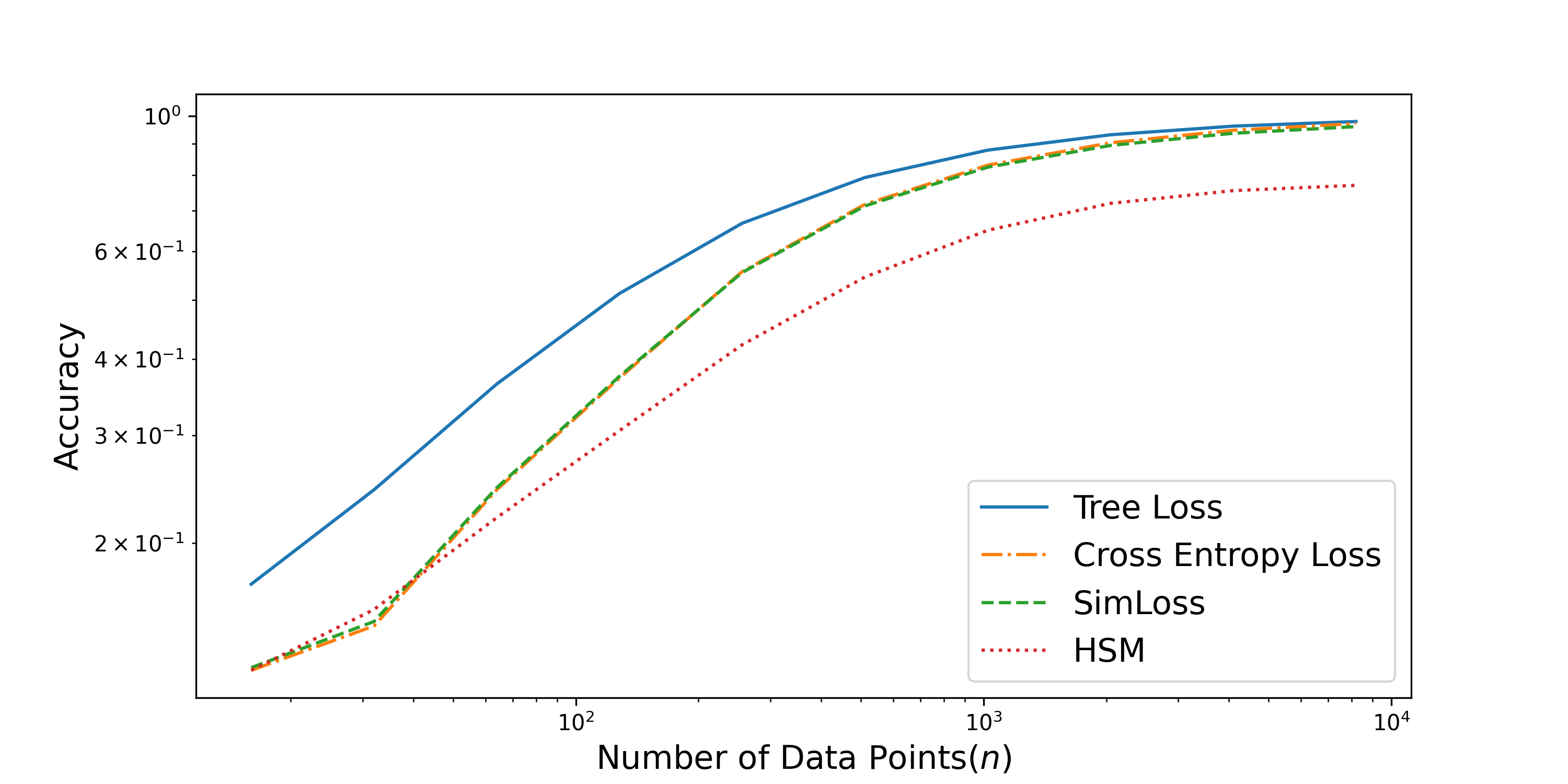}
\includegraphics[width=\columnwidth,height=1.45in]{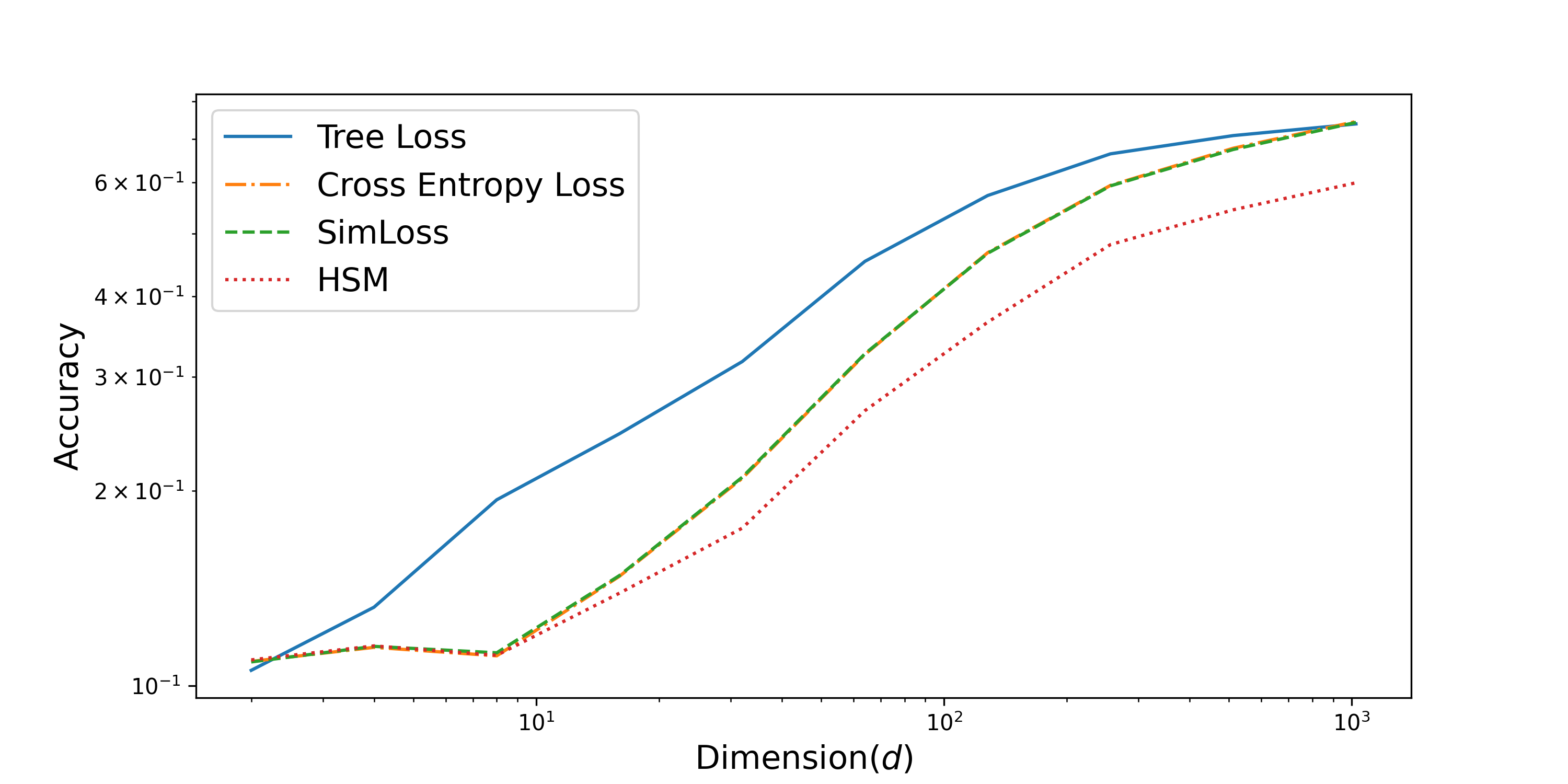}

\includegraphics[width=\columnwidth,height=1.45in]{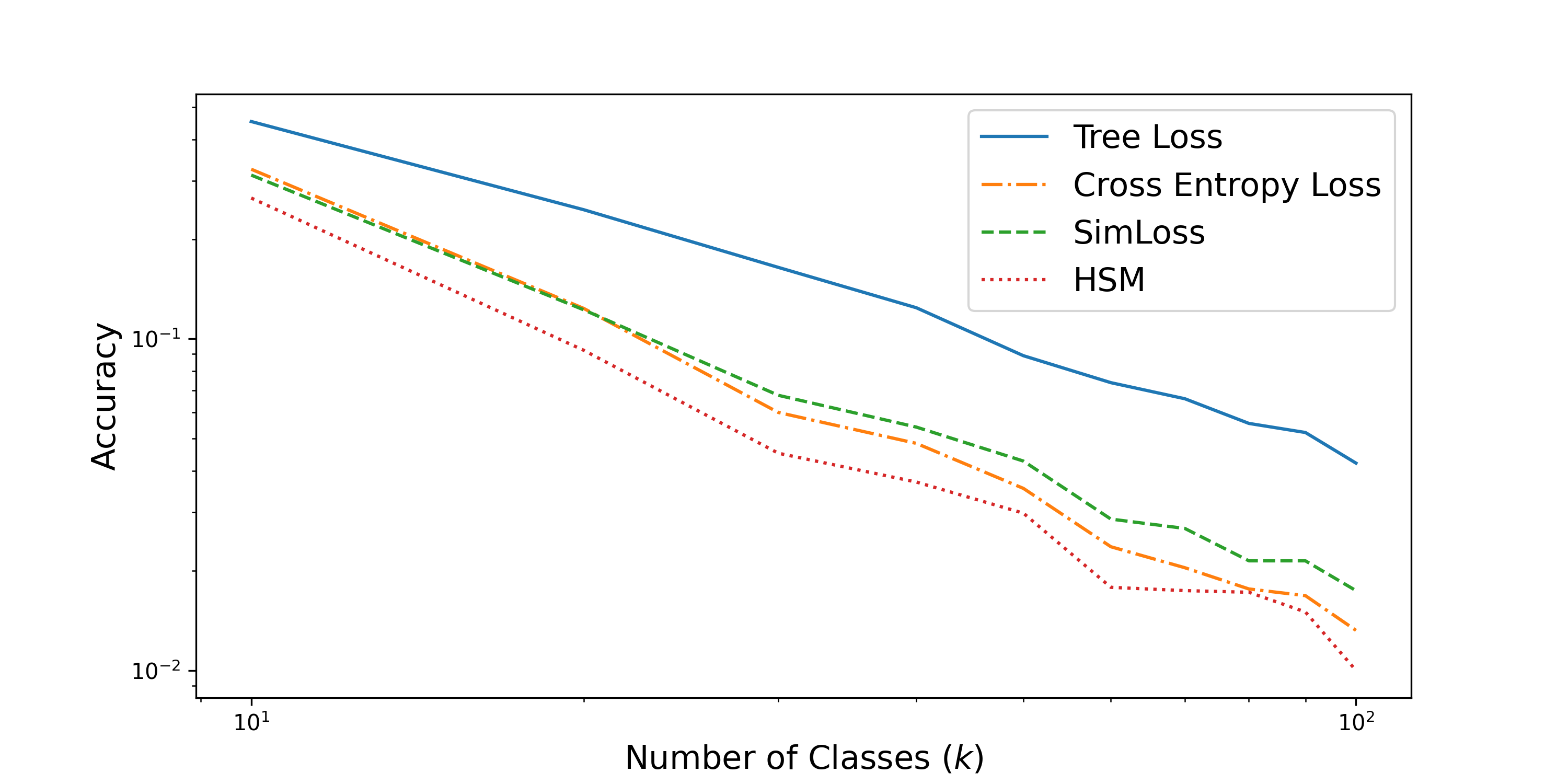}
\includegraphics[width=\columnwidth,height=1.45in]{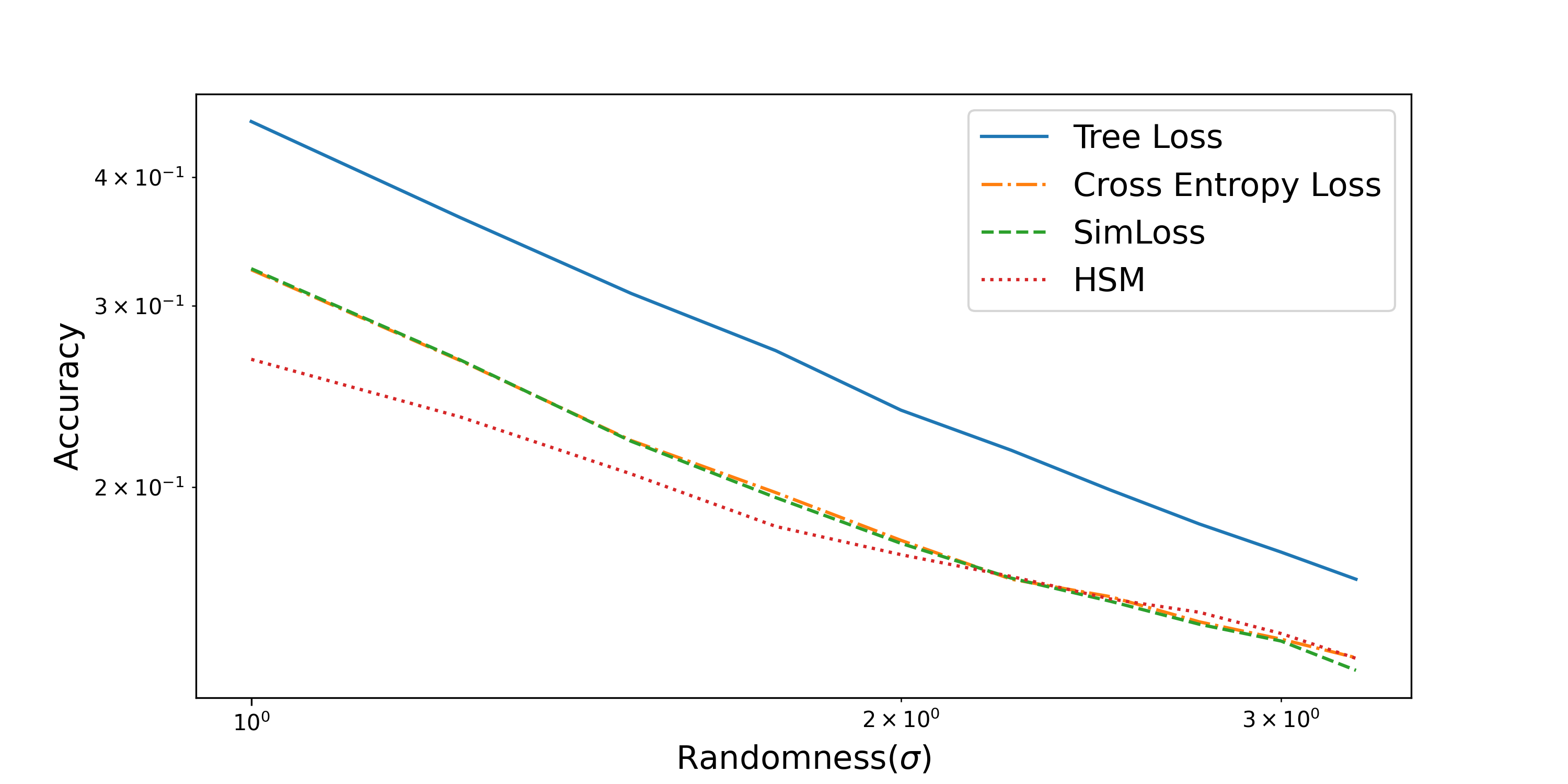}
\caption{
    Results of Synthetic Experiment I.
    The x-axis of each plot shows which problem parameter is being varied.
    As our theory predicts, the Tree Loss outperforms the baseline loss functions in all data regimes.
}
\label{fig:synth:1}
\end{figure*}

\section{Experiments}
\label{sec:experiment}

We evaluate the tree loss on 4 synthetic and 3 real world experiments.
We compare the tree loss to the standard cross entropy loss, the recently proposed SimLoss \citep{Kobs2020SimLossCS}, and the hierarchical softmax \citep{morin2005hierarchical}.
The results confirm our theoretical findings from Section \ref{sec:theory} and demonstrate that the tree loss significantly improves the performance of these baseline losses in a wide range of scenarios.

\subsection{Synthetic data}

Experiments on synthetic data let us control various hyperparameters of the dataset in order to see how the tree loss behaves in a wide range of scenarios.
The next subsection introduces our data generation procedure,
and then subsequent subsections describe 4 different experiments baed on this procedure.

\subsubsection{Data Generation Procedure}
\label{sec:exp:synth:problem}

Our data generation procedure has 4 key hyperparameters:
the number of data points $n$, the number of feature dimensions $d$, the number of classes $k$, and a randomness parameter $\sigma$.

Let $\mathcal N$ be the standard normal distribution.
Then sample the true parameter matrix as
\begin{align}
    \star W \sim \mathcal N^{k\times d}
    .
\end{align}
Standard results on random matrices show that $\lF{\star W} = O(\sqrt{kd})$ with high probability, as assumed by our theory.

For each data point $i\in[n]$,
we sample the data point according to the following rules:
\begin{align}
    y_i &\sim \text{Uniform}([k]), \text{and} \\
    \x_{i} &\sim \mathcal N(\w^*_{y_i}; \sigma).
\end{align}
Observe that larger $\sigma$ values result in
more noise in the data points,
making the classes harder to distinguish,
and increasing the bayes error of the problem.
Also observe that for any two classes $i$ and $j$,
the distance between the two classes $\ltwo{\star\w_i - \star\w_j} = O(\sqrt{d})$.
This implies that as $d$ increases,
the separation between data points from different classes will also increase,
reducing the bayes error of the problem.


\subsubsection{Experiment I: Data Regimes}
\label{sec:synth:1}

Our first experiment investigates the tree loss's performance in a wide range of statistical settings controlled by the dataset hyperparameters ($n$, $d$, $k$, and $\sigma$).
We use a baseline experimental setting with $n=100$, $d=64$, $k=10$, and $\sigma=1$.
For each of the hyperparameters, we investigate the effect of that hyperparameter's performance on the problem by varying it over a large range.
For each value in the range, we:
(1) randomly sample 50 different $\star W$ using the procedure described in Section \ref{sec:exp:synth:problem};
(2) for each $\star W$, we:
(2a) sample a training set with $n$ data points and a test set with 10000 data points\footnote{The size of the test set is fixed and large to ensure that our results have high statistical significance.};
(2b) train a separate model for each of our loss functions;
(2c) report the average test accuracy across 50 samples from step (1).
Figure \ref{fig:synth:1} shows the results of these experiments.
As our theory suggests, the tree loss outperforms all other losses in all regimes.

Consider the top-left plot.
This plot has a low bayes error, and so all the variation in performance we see is due to the generalization error of the models.
As the number of data points $n$ grows large,
its influence on the standard cross entropy's convergence rate of $O(\sqrt{k/n})$ dominates,
and the model generalizes well.
When the number of data points is small, 
then the dependence on $k$ becomes more important.
The tree loss's dependence of $O(\sqrt{k^{1-1/c}/n})$ is strictly better than the cross entropy loss's,
explaining the observed improved performance.
A similar effect explain's our model's observed improved performance in the bottom-left plot as $k$ varies.

Now consider the top-right plot.
The bayes error of our problem setup is inversely related to the problem dimension $d$ (see observations in Section \ref{sec:exp:synth:problem}),
so this plot compares performance on different problem difficulties.
On maximally easy (large $d$) and maximally hard (small $d$) problems,
the tree loss performs similarly to the other loss functions because the performance is dominated by the bayes error and no improvement is possible.
The main advantage of the tree loss is in the mid-difficulty problems.
In this regime, performance is dominated by the generalization ability of the different losses,
and the tree loss's improved generalization results in noticeable improvements.
The bottom-right plot tells a similar story but controls the difficulty of the problem directly by tuning the randomness $\sigma$.

\subsubsection{Experiment II: Parameter Norms}

\begin{figure}
\includegraphics[width=\columnwidth,height=1.5in]{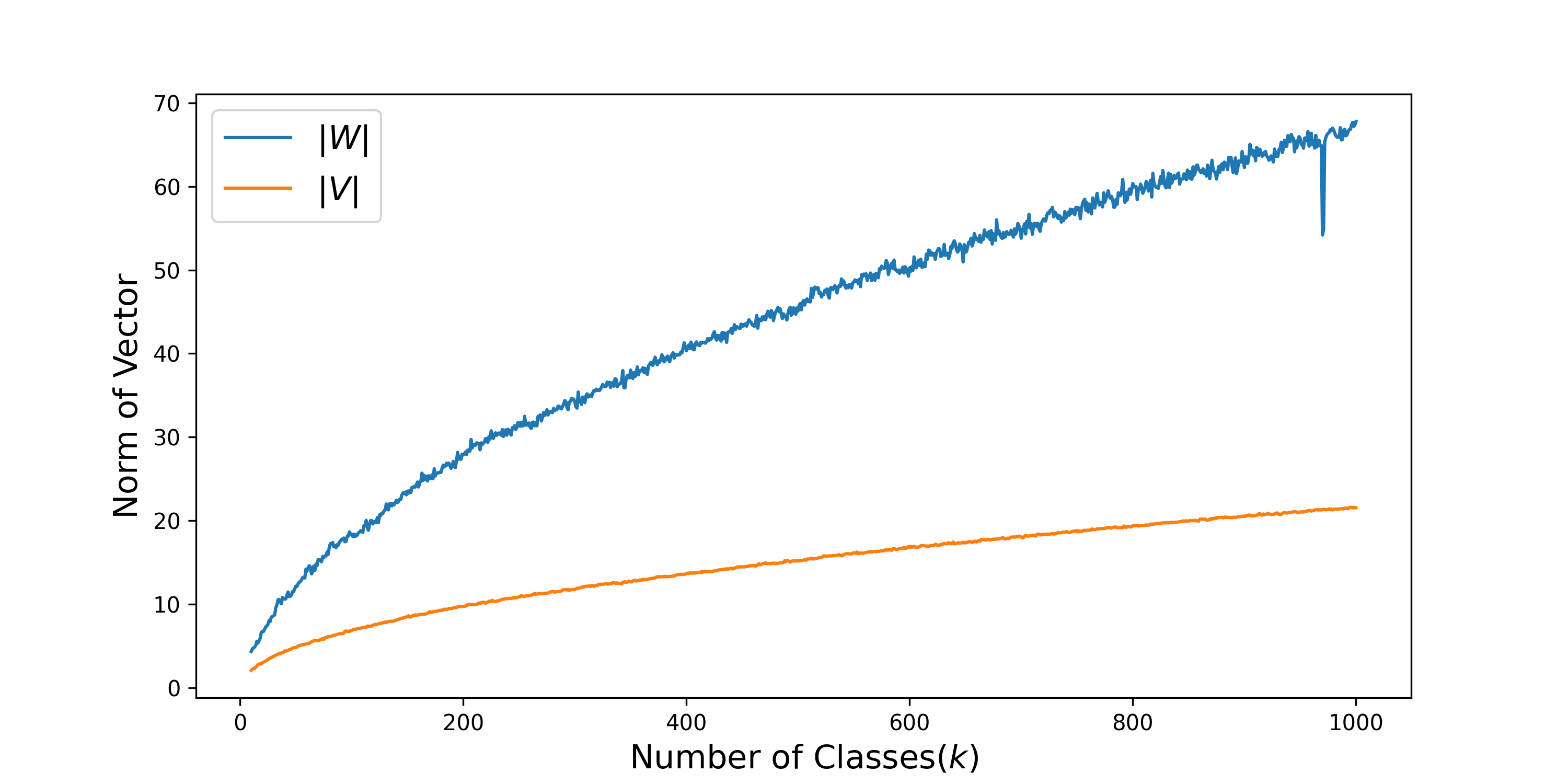}
\caption{
    Results of Synthetic Experiment II.
    As we increase $k$,
    $\lF{\star V}$ grows significantly slower than $\lF{\star W}$.
    Since the convergence rate of SGD is proportional to $\lF{\cdot}$ (by Lemma \ref{ref:cor:A}),
    the tree loss converges faster than the standard cross entropy loss. 
    }
\label{fig:synth:norm}
\end{figure}

Lemma \ref{ref:cor:A} suggests that the norm of the parameter matrix is what controls convergence rate of SGD,
and the proof of our main result in Corollary \ref{cor:main} relies on bounding this norm.
In this experiment, we directly measure the norm of the parameter matrices and show that $\lF{V}$ is significantly less than $\lF{W}$,
justifying the good convergence rates we observed in Synthetic Experiment I above.
We set $n=1000$, $d=10$, and $\sigma=1$, and vary the number of classes $k$.
Figure \ref{fig:synth:norm} shows that $\lF{\star V}$ grows much slower than $\lF{\star W}$.
Notice in particular that $\lF{\star W}$ grows at the rate $O(\sqrt{k})$ and $\lF{\star V}$ grows at a rate strictly less than $o(\sqrt{k})$ as our theory predicts.

\subsubsection{Experiment III: Tree Shapes}

In this experiment, we study how the shape of our tree structure impacts performance.
We set $n=1000$, $d=10$, $k=100$, and $\sigma=1$.
We use a relatively large number of classes $k$ compared to the standard problem in Section \ref{sec:synth:1} above in order to ensure that we have enough classes to generate meaningfully complex tree structures.
The basic trends we observe are consistent for all other values of $n$, $d$, and $\sigma$.

\begin{figure}
\includegraphics[width=\columnwidth,height=1.5in]{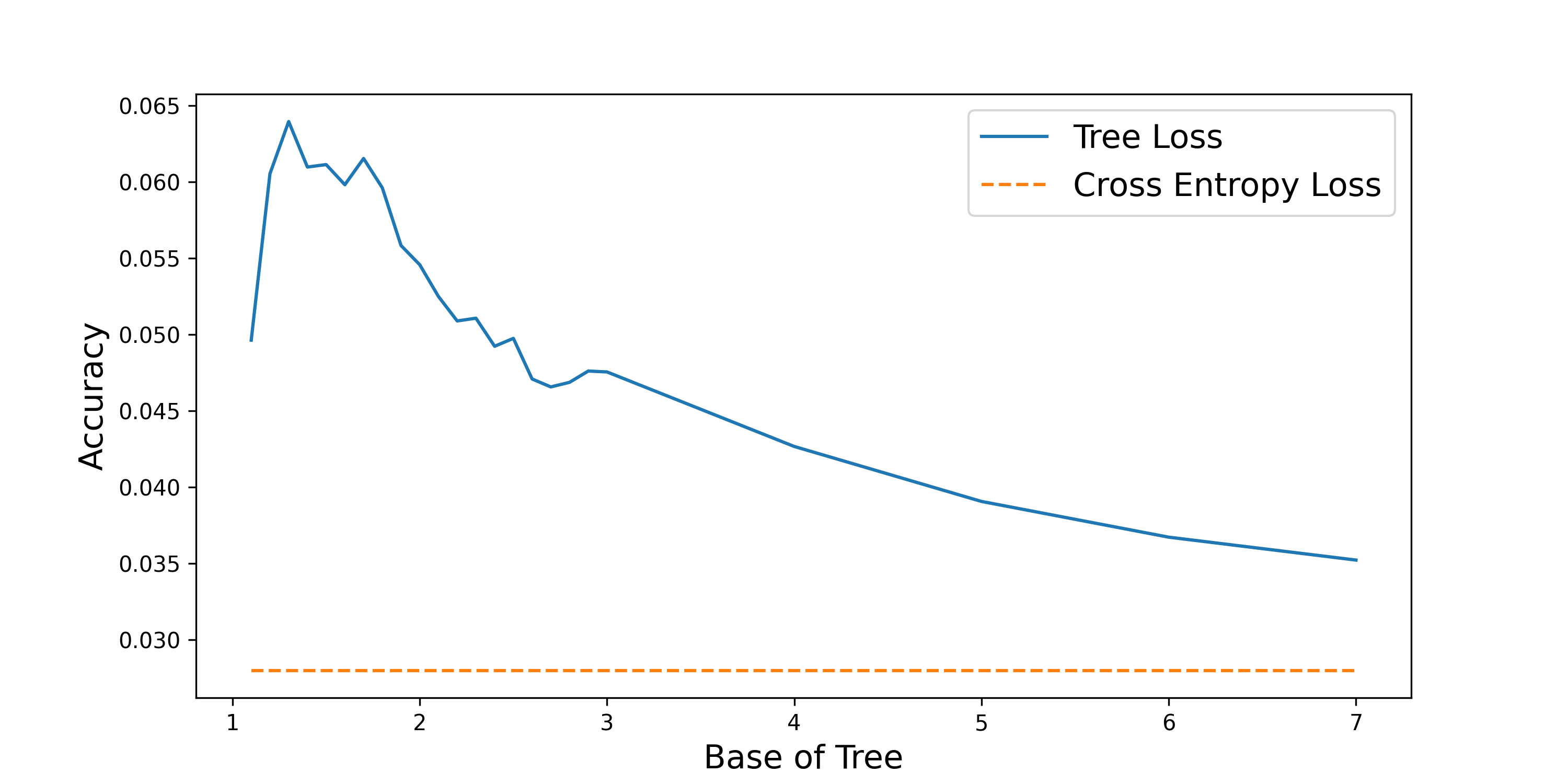}
\caption
{
    Results of Synthetic Experiment III.
    The tree loss outperforms the standard cross entropy loss for all values of \texttt{base}.
}
\label{fig:ct:acc}
\end{figure}

Recall that the cover tree has a parameter \texttt{base} which controls the rate of expansion between layers of the tree.
Reducing \texttt{base} increases the height of the tree and increasing \texttt{base} reduces the height of the tree.
This will affect the performance of our tree loss because taller trees result in more parameter sharing.
Figure \ref{fig:ct:acc} plots the accuracy of the tree loss as a function of the \texttt{base} hyperparameter.
Across all ranges, the tree loss outperforms the cross entropy loss.
Interestingly, the tree loss's accuracy is maximized when $\texttt{base}\approx1.3$.
The original cover tree paper \citep{beygelzimer2006cover} also found that a \texttt{base} of 1.3 resulted in the fastest nearest neighbor queries.
We do not know of a good theoretical explanation for this phenomenon.


\subsubsection{Experiment IV: Metric Quality}
\label{sec:synth:eps}

Recall that our theoretical results state that the convergence rate of SGD depends on a factor $\lambda$ that measures how well our metric over the class labels is able to predict the distances between the true parameter vectors.
(See Assumption \ref{ass:metric}.)
In this experiment, we study the effect of this $\lambda$ parameter on prediction accuracy.

We fix the following problem hyperparameters:
$n=100$, $d=64$, $k=10$, and $\sigma=1$.
Then we construct a bad parameter matrix $\bad W$ using the same procedure we used to construct the optimal parameter matrix;
that is,
\begin{equation}
    \bad W \sim \mathcal N ^ {k\times d}
    .
\end{equation}
Each row $\bad \w_i$ of $\bad W$ is now a $d$ dimensional random vector that has absolutely no relation to the true parameter vector.
Next, we define a family of ``$\epsilon$-bad'' parameter vectors that mix between the bad and optimal parameter vectors:
\begin{equation}
    \w^\epsilon_i = (1-\epsilon) \w^*_i + \epsilon \bad\w_i.
\end{equation}
Finally, we define our $\epsilon$-bad distance metric as
\begin{equation}
    d_\epsilon(i,j) = \ltwo{\w_i^\epsilon - \w_j^\epsilon}
\end{equation}
and build our cover tree structure using $d_\epsilon$.
When $\epsilon=0$, this cover tree structure will be the ideal structure and $\lambda$ will be equal to 1;
when $\epsilon=1$, this cover tree structure will be maximally bad, and $\lambda$ will be large.

\begin{figure}
\includegraphics[width=\columnwidth,height=1.5in]{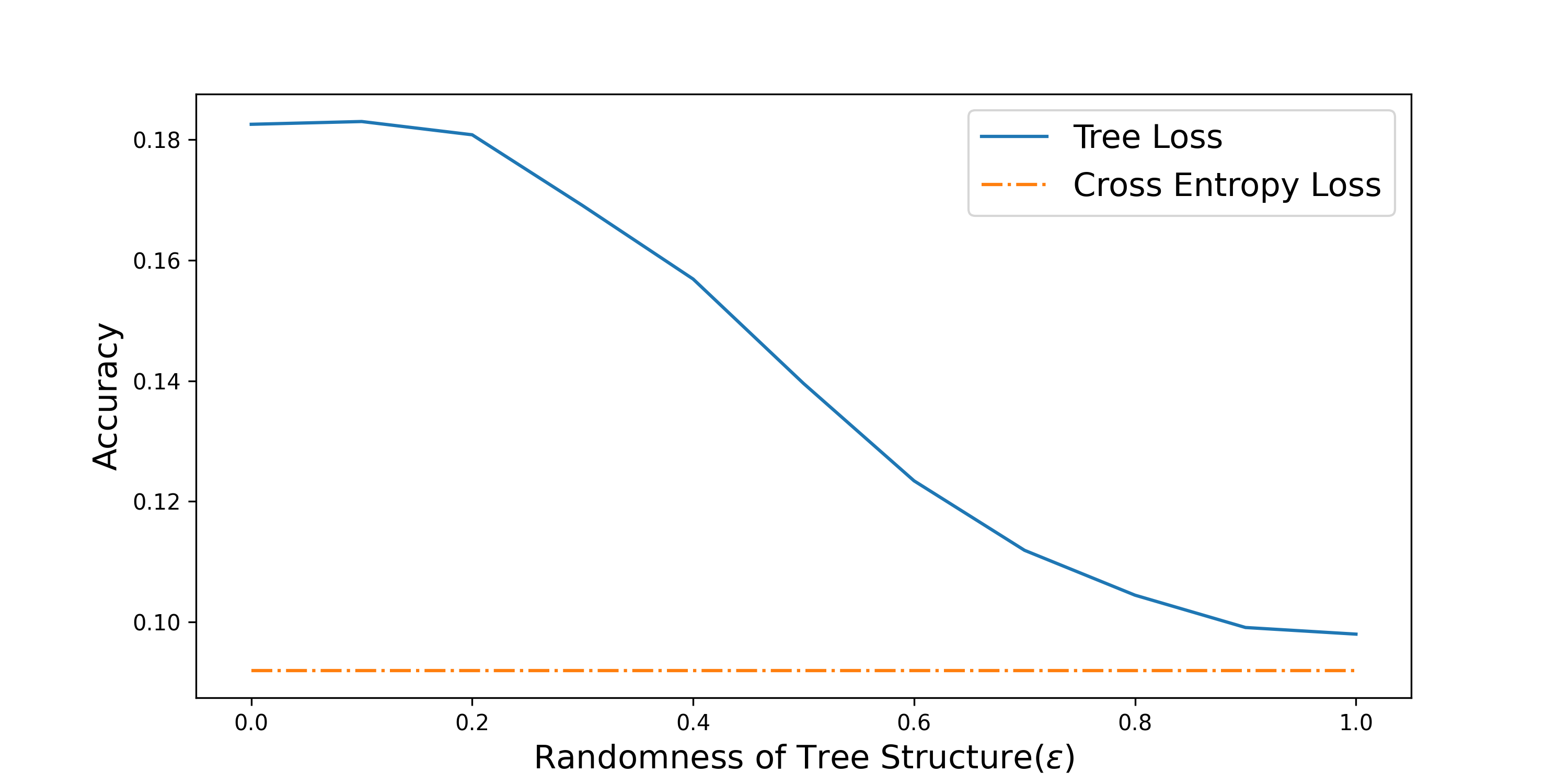}
\caption{
Results for Synthetic Experiment IV.
As $\epsilon\to1$, 
the tree structure becomes perfectly random.
The tree loss still outperforms the cross entropy loss in this worst-case scenario.
}  
\label{fig:synth:eps}
\end{figure}

Figure \ref{fig:synth:eps} shows the performance of the tree loss as a function of $\epsilon$.
Remarkably, the tree loss outperforms the standard cross entropy loss even when using a perfectly bad tree structure (i.e. when $\epsilon=1$).
This surprising empirical finding actually agrees with our theoretical results in two ways.
First, Lemma \ref{lemma:starU} implies that that $\lF{\star V} = O(\lF{\star W})$,
and so no tree structure can perform asymptotically worse than the standard cross entropy loss.
Second, when $\epsilon=1$ and we have a perfectly random tree structure,
the distance $d_\epsilon$ can still be embedded into the ideal metric $d$ with some large (but finite!) $\lambda$.
Asymptotically, Corollary \ref{cor:main} then implies that SGD converges at the rate $O(\sqrt{k^{1-1/c}})$,
which is less than the convergence rate of the standard cross entropy loss.

\subsection{Real World Data}

\begin{table*}
    \centering
    \begin{tabular}{@{}lR{0.75in}R{0.75in}R{0.75in}R{0.75in}R{0.75in}R{0.75in}@{}}
                      & \multicolumn{2}{c}{CIFAR100}                           & \multicolumn{2}{c}{ImageNet}          & \multicolumn{2}{c}{Twitter} \\
\cmidrule(l){2-3} 
\cmidrule(l){4-5} 
\cmidrule(l){6-7} 
                      & Top1 Acc        & SA                & Top1 Acc          & SA         & Top1 Acc          & SA \\
\midrule
tree loss             & \textbf{52.78}  & \textbf{65.90}    & \textbf{76.52}    & \textbf{88.22}    & \textbf{20.96}    & \textbf{50.54} \\
cross entropy loss    & 51.87           & 65.58             & 75.30             & 87.51             & 19.95             & 48.33 \\
simloss               & 52.43           & 65.82             & 76.24             & 87.98             & 19.96             & 48.32 \\
\bottomrule
\end{tabular}

    \caption{Experimental results on real world datasets.  For all performance measures, larger numbers are better.  The tree loss achieves the best results in all cases.}
    \label{table:results}
\end{table*}

We now validate the tree loss on real world image (CIFAR100, ImageNet) and text (Twitter) data.
For each dataset, we report standard top-1 accuracy scores,
and the less well known \emph{similarity accuracy} (SA) score.
SA is a top-1 accuracy considering similarity.
The idea is that misclassifying a \texttt{sheepdog} as a \texttt{husky} should be penalized less than misclassifying a \texttt{sheepdog} as a \texttt{bus} because the \texttt{sheepdog} and \texttt{husky} classes are more similar to each other.

Table \ref{table:results} shows that on each of these datasets and for each metric,
the tree loss outperforms the baseline cross entropy loss and SimLoss.\footnote{We do not evaluate against the hierarchical softmax due to limited computational resources.  The results on the synthetic data, however, suggest that the hierarchical softmax would have performed poorly.}
The remainder of this section describes the experimental procedures for each dataset in detail.

\subsubsection{CIFAR100}

CIFAR100 is a standard image dataset with 100 classes \citep{krizhevsky2009learning}.
It is more difficult than MNIST and CIFAR10,
but small and so less computationally demanding than ImageNet.
In this experiment, we exactly follow the procedure used in the SimLoss paper \citep{Kobs2020SimLossCS}, 
and find that under their experimental conditions,
the tree loss has the best performance.

First, we generate our distance metric over the class labels using a word2vec model \citep{Mikolov2013EfficientEO} pretrained on GoogleNews.
The distance between class labels is defined to be the distance between their corresponding word2vec embeddings.
There are 4 class labels with no corresponding word2vec embedding,
and so following the SimLoss paper we discard these classes.

We train 3 ResNet20 models on CIFAR100 \citep{He2016DeepRL} using hyperparameters specified in the original paper.
The only difference between the three models is the final layer:
one model uses the standard cross entropy loss, one uses the tree loss, and one uses the SimLoss.
The results shown in Table \ref{table:results} shows that the tree loss significantly outperforms the other losses.

\subsubsection{ImageNet}

ImageNet is the gold-standard dataset for image classification tasks \citep{Russakovsky2015ImageNetLS}.
It has 1000 class labels and 1.2 million images.

We generate our distance metric over the image labels using a pretrained fastText model \citep{bojanowski2017enriching}.
We use fastText rather than word2vec because many of the class labels contain words not in the word2vec model's vocabulary;
since fastText is naturally able to handle out-of-vocabulary words,
there is no need to discard class labels like we did for CIFAR100.
Many ImageNet class label names contain multiple words,
and for these classes we generate the embedding with a simple average of the fastText embedding over all words.

We again train a ResNet50 model \citep{He2016DeepRL} using standard hyperparameters, replacing only the last layer of the model.
As with CIFAR100, Table \ref{table:results} shows that the tree loss significantly outperforms the other losses.
Since the publication of ResNet50 in 2016,
there have been many newer network architectures with better performance on ImageNet \citep[e.g.][]{howard2017mobilenets,huang2017densely,pmlr-v97-tan19a}.
We unfortunately did not have sufficient computational resources to train these more modern models with the tree loss,
and we emphasize that we are not trying to compete directly with these network architectures to achieve state-of-the-art performance.
Instead, our goal is to show that for a given network architecture,
replacing the cross entropy loss with the tree loss results in improved performance.

\subsubsection{Twitter Data}

We now consider the problem of predicting the emotions of Twitter text.
This is a very different domain than the image problems considered above,
and demonstrates that the tree loss works across many domains.
We use the dataset collected by \citet{izbicki2019geolocating} and \citet{stoikos2020multilingual}.
This dataset contains all geolocated tweets written in more than 100 languages sent over the 4 year period between October 2017 and October 2021---approximately $5.5\times10^9$ tweets in total.
Tweets in the dataset are preprocessed to have all emojis, usernames, and urls removed from the text,
and then the goal is to predict the emoji given only the remaining text.
Since most emojis represent emotions,
the task of emoji prediction serves as a proxy for emotion prediction.

We generate our distance metric over the class labels using the pretrained emoji2vec model \citep{Eisner2016emoji2vecLE},
which associates each emoji with a vector embedding.
We then follow the procedure in \citet{stoikos2020multilingual} to train multilingual BERT models \citep{Feng2020LanguageagnosticBS} with the last layers replaced by the three different loss functions.
Table \ref{table:results} shows the tree loss significantly outperforming the baseline models.

\section{Conclusion}
The tree loss is a drop-in replacement for the cross entropy loss for multiclass classification problems.
It can take advantage of background knowledge about the underlying class structure in order to improve the convergence rate of SGD.
Both theoretical and empirical results suggest there is no disadvantage for using the tree loss,
but potentially large advantages.


\bibliographystyle{plainnat}
\bibliography{paper}

\begin{thebibliography}{31}
\providecommand{\natexlab}[1]{#1}
\providecommand{\url}[1]{\texttt{#1}}
\expandafter\ifx\csname urlstyle\endcsname\relax
  \providecommand{\doi}[1]{doi: #1}\else
  \providecommand{\doi}{doi: \begingroup \urlstyle{rm}\Url}\fi

\bibitem[Abadi et~al.(2015)Abadi, Agarwal, Barham, Brevdo, Chen, Citro,
  Corrado, Davis, Dean, Devin, Ghemawat, Goodfellow, Harp, Irving, Isard, Jia,
  Jozefowicz, Kaiser, Kudlur, Levenberg, Man\'{e}, Monga, Moore, Murray, Olah,
  Schuster, Shlens, Steiner, Sutskever, Talwar, Tucker, Vanhoucke, Vasudevan,
  Vi\'{e}gas, Vinyals, Warden, Wattenberg, Wicke, Yu, and
  Zheng]{tensorflow2015-whitepaper}
Mart\'{\i}n Abadi, Ashish Agarwal, Paul Barham, Eugene Brevdo, Zhifeng Chen,
  Craig Citro, Greg~S. Corrado, Andy Davis, Jeffrey Dean, Matthieu Devin,
  Sanjay Ghemawat, Ian Goodfellow, Andrew Harp, Geoffrey Irving, Michael Isard,
  Yangqing Jia, Rafal Jozefowicz, Lukasz Kaiser, Manjunath Kudlur, Josh
  Levenberg, Dandelion Man\'{e}, Rajat Monga, Sherry Moore, Derek Murray, Chris
  Olah, Mike Schuster, Jonathon Shlens, Benoit Steiner, Ilya Sutskever, Kunal
  Talwar, Paul Tucker, Vincent Vanhoucke, Vijay Vasudevan, Fernanda Vi\'{e}gas,
  Oriol Vinyals, Pete Warden, Martin Wattenberg, Martin Wicke, Yuan Yu, and
  Xiaoqiang Zheng.
\newblock {TensorFlow}: Large-scale machine learning on heterogeneous systems,
  2015.
\newblock URL \url{https://www.tensorflow.org/}.
\newblock Software available from tensorflow.org.

\bibitem[Bertinetto et~al.(2020)Bertinetto, Mueller, Tertikas, Samangooei, and
  Lord]{bertinetto2020making}
Luca Bertinetto, Romain Mueller, Konstantinos Tertikas, Sina Samangooei, and
  Nicholas~A Lord.
\newblock Making better mistakes: Leveraging class hierarchies with deep
  networks.
\newblock In \emph{Proceedings of the IEEE/CVF Conference on Computer Vision
  and Pattern Recognition}, pages 12506--12515, 2020.

\bibitem[Beygelzimer et~al.(2006)Beygelzimer, Kakade, and
  Langford]{beygelzimer2006cover}
Alina Beygelzimer, Sham Kakade, and John Langford.
\newblock Cover trees for nearest neighbor.
\newblock In \emph{Proceedings of the 23rd international conference on Machine
  learning}, pages 97--104, 2006.

\bibitem[Bojanowski et~al.(2017)Bojanowski, Grave, Joulin, and
  Mikolov]{bojanowski2017enriching}
Piotr Bojanowski, Edouard Grave, Armand Joulin, and Tomas Mikolov.
\newblock Enriching word vectors with subword information.
\newblock \emph{Transactions of the Association for Computational Linguistics},
  5:\penalty0 135--146, 2017.

\bibitem[Cesa-Bianchi et~al.(2006)Cesa-Bianchi, Gentile, and
  Zaniboni]{cesa2006incremental}
Nicolo Cesa-Bianchi, Claudio Gentile, and Luca Zaniboni.
\newblock Incremental algorithms for hierarchical classification.
\newblock \emph{The Journal of Machine Learning Research}, 7:\penalty0 31--54,
  2006.

\bibitem[Eisner et~al.(2016)Eisner, Rockt{\"a}schel, Augenstein, Bosnjak, and
  Riedel]{Eisner2016emoji2vecLE}
Ben Eisner, Tim Rockt{\"a}schel, Isabelle Augenstein, Matko Bosnjak, and
  S.~Riedel.
\newblock emoji2vec: Learning emoji representations from their description.
\newblock \emph{ArXiv}, abs/1609.08359, 2016.

\bibitem[Feng et~al.(2020)Feng, Yang, Cer, Arivazhagan, and
  Wang]{Feng2020LanguageagnosticBS}
Fangxiaoyu Feng, Yin-Fei Yang, Daniel~Matthew Cer, N.~Arivazhagan, and Wei
  Wang.
\newblock Language-agnostic bert sentence embedding.
\newblock \emph{ArXiv}, abs/2007.01852, 2020.

\bibitem[He et~al.(2016)He, Zhang, Ren, and Sun]{He2016DeepRL}
Kaiming He, X.~Zhang, Shaoqing Ren, and Jian Sun.
\newblock Deep residual learning for image recognition.
\newblock \emph{2016 IEEE Conference on Computer Vision and Pattern Recognition
  (CVPR)}, pages 770--778, 2016.

\bibitem[Howard et~al.(2017)Howard, Zhu, Chen, Kalenichenko, Wang, Weyand,
  Andreetto, and Adam]{howard2017mobilenets}
Andrew~G Howard, Menglong Zhu, Bo~Chen, Dmitry Kalenichenko, Weijun Wang,
  Tobias Weyand, Marco Andreetto, and Hartwig Adam.
\newblock Mobilenets: Efficient convolutional neural networks for mobile vision
  applications.
\newblock \emph{arXiv preprint arXiv:1704.04861}, 2017.

\bibitem[Huang et~al.(2017)Huang, Liu, Van Der~Maaten, and
  Weinberger]{huang2017densely}
Gao Huang, Zhuang Liu, Laurens Van Der~Maaten, and Kilian~Q Weinberger.
\newblock Densely connected convolutional networks.
\newblock In \emph{Proceedings of the IEEE conference on computer vision and
  pattern recognition}, pages 4700--4708, 2017.

\bibitem[Izbicki(2017)]{izbickithesis}
Mike Izbicki.
\newblock \emph{Divide and Conquer Algorithms for Machine Learning}.
\newblock PhD thesis, University of California, Riverside, 2017.

\bibitem[Izbicki and Shelton(2015)]{izbicki2015faster}
Mike Izbicki and Christian Shelton.
\newblock Faster cover trees.
\newblock In \emph{International Conference on Machine Learning}, pages
  1162--1170. PMLR, 2015.

\bibitem[Izbicki et~al.(2019)Izbicki, Papalexakis, and
  Tsotras]{izbicki2019geolocating}
Mike Izbicki, Vagelis Papalexakis, and Vassilis Tsotras.
\newblock Geolocating tweets in any language at any location.
\newblock In \emph{Proceedings of the 28th ACM International Conference on
  Information and Knowledge Management}, pages 89--98, 2019.

\bibitem[Jiang et~al.(2017)Jiang, Rong, Gao, Shen, and
  Xiong]{Jiang2017ExplorationOT}
Nan Jiang, Wenge Rong, M.~Gao, Yikang Shen, and Z.~Xiong.
\newblock Exploration of tree-based hierarchical softmax for recurrent language
  models.
\newblock In \emph{IJCAI}, 2017.

\bibitem[Kobs et~al.(2020)Kobs, Steininger, Zehe, Lautenschlager, and
  Hotho]{Kobs2020SimLossCS}
Konstantin Kobs, M.~Steininger, Albin Zehe, Florian Lautenschlager, and
  A.~Hotho.
\newblock Simloss: Class similarities in cross entropy.
\newblock \emph{ArXiv}, abs/2003.03182, 2020.

\bibitem[Krizhevsky et~al.(2009)Krizhevsky, Hinton,
  et~al.]{krizhevsky2009learning}
Alex Krizhevsky, Geoffrey Hinton, et~al.
\newblock Learning multiple layers of features from tiny images.
\newblock 2009.

\bibitem[Lapin et~al.(2016)Lapin, Hein, and Schiele]{lapin2016loss}
Maksim Lapin, Matthias Hein, and Bernt Schiele.
\newblock Loss functions for top-k error: Analysis and insights.
\newblock In \emph{Proceedings of the IEEE Conference on Computer Vision and
  Pattern Recognition}, pages 1468--1477, 2016.

\bibitem[Mikolov et~al.(2013)Mikolov, Chen, Corrado, and
  Dean]{Mikolov2013EfficientEO}
Tomas Mikolov, Kai Chen, G.~Corrado, and J.~Dean.
\newblock Efficient estimation of word representations in vector space.
\newblock In \emph{ICLR}, 2013.

\bibitem[Mohammed and Umaashankar(2018)]{Mohammed2018EffectivenessOH}
Abdul~Arfat Mohammed and Venkatesh Umaashankar.
\newblock Effectiveness of hierarchical softmax in large scale classification
  tasks.
\newblock \emph{2018 International Conference on Advances in Computing,
  Communications and Informatics (ICACCI)}, pages 1090--1094, 2018.

\bibitem[Morin and Bengio(2005)]{morin2005hierarchical}
Frederic Morin and Yoshua Bengio.
\newblock Hierarchical probabilistic neural network language model.
\newblock In \emph{Aistats}, volume~5, pages 246--252. Citeseer, 2005.

\bibitem[Paszke et~al.(2019)Paszke, Gross, Massa, Lerer, Bradbury, Chanan,
  Killeen, Lin, Gimelshein, Antiga, Desmaison, Kopf, Yang, DeVito, Raison,
  Tejani, Chilamkurthy, Steiner, Fang, Bai, and Chintala]{NEURIPS2019_9015}
Adam Paszke, Sam Gross, Francisco Massa, Adam Lerer, James Bradbury, Gregory
  Chanan, Trevor Killeen, Zeming Lin, Natalia Gimelshein, Luca Antiga, Alban
  Desmaison, Andreas Kopf, Edward Yang, Zachary DeVito, Martin Raison, Alykhan
  Tejani, Sasank Chilamkurthy, Benoit Steiner, Lu~Fang, Junjie Bai, and Soumith
  Chintala.
\newblock Pytorch: An imperative style, high-performance deep learning library.
\newblock In \emph{Advances in Neural Information Processing Systems 32}, pages
  8024--8035. 2019.

\bibitem[Peng et~al.(2017)Peng, Li, Song, and Liu]{Peng2017IncrementallyLT}
Hao Peng, Jianxin Li, Y.~Song, and Yaopeng Liu.
\newblock Incrementally learning the hierarchical softmax function for neural
  language models.
\newblock In \emph{AAAI}, 2017.

\bibitem[Rakhlin et~al.(2012)Rakhlin, Shamir, and Sridharan]{rakhlin2011making}
Alexander Rakhlin, Ohad Shamir, and Karthik Sridharan.
\newblock Making gradient descent optimal for strongly convex stochastic
  optimization.
\newblock \emph{ICML}, 2012.

\bibitem[Russakovsky et~al.(2015)Russakovsky, Deng, Su, Krause, Satheesh, Ma,
  Huang, Karpathy, Khosla, Bernstein, Berg, and
  Fei-Fei]{Russakovsky2015ImageNetLS}
Olga Russakovsky, J.~Deng, Hao Su, J.~Krause, S.~Satheesh, S.~Ma, Zhiheng
  Huang, A.~Karpathy, A.~Khosla, Michael~S. Bernstein, A.~Berg, and Li~Fei-Fei.
\newblock Imagenet large scale visual recognition challenge.
\newblock \emph{International Journal of Computer Vision}, 115:\penalty0
  211--252, 2015.

\bibitem[Shalev-Shwartz and Ben-David(2014)]{shalev2014understanding}
Shai Shalev-Shwartz and Shai Ben-David.
\newblock \emph{Understanding machine learning: From theory to algorithms}.
\newblock Cambridge university press, 2014.

\bibitem[Stoikos and Izbicki(2020)]{stoikos2020multilingual}
Stefanos Stoikos and Mike Izbicki.
\newblock Multilingual emoticon prediction of tweets about covid-19.
\newblock In \emph{Proceedings of the Third Workshop on Computational Modeling
  of People's Opinions, Personality, and Emotion's in Social Media}, pages
  109--118, 2020.

\bibitem[Sukhbaatar et~al.(2015)Sukhbaatar, Bruna, Paluri, Bourdev, and
  Fergus]{sukhbaatar2014training}
Sainbayar Sukhbaatar, Joan Bruna, Manohar Paluri, Lubomir Bourdev, and Rob
  Fergus.
\newblock Training convolutional networks with noisy labels.
\newblock \emph{ICLR}, 2015.

\bibitem[Tan and Le(2019)]{pmlr-v97-tan19a}
Mingxing Tan and Quoc Le.
\newblock {E}fficient{N}et: Rethinking model scaling for convolutional neural
  networks.
\newblock In Kamalika Chaudhuri and Ruslan Salakhutdinov, editors,
  \emph{Proceedings of the 36th International Conference on Machine Learning},
  volume~97 of \emph{Proceedings of Machine Learning Research}, pages
  6105--6114. PMLR, 09--15 Jun 2019.
\newblock URL \url{https://proceedings.mlr.press/v97/tan19a.html}.

\bibitem[Wu et~al.(2019)Wu, Tygert, and LeCun]{wu2017hierarchical}
Cinna Wu, Mark Tygert, and Yann LeCun.
\newblock A hierarchical loss and its problems when classifying
  non-hierarchically.
\newblock \emph{PLoS ONE}, 2019.

\bibitem[Yang et~al.(2017)Yang, Ruan, Li, and Hu]{Yang2017OptimizeHS}
Zhixuan Yang, Chong Ruan, Caihua Li, and J.~Hu.
\newblock Optimize hierarchical softmax with word similarity knowledge.
\newblock \emph{Polytech. Open Libr. Int. Bull. Inf. Technol. Sci.},
  55:\penalty0 11--16, 2017.

\bibitem[Zhang and Sabuncu(2018)]{zhang2018generalized}
Zhilu Zhang and Mert~R Sabuncu.
\newblock Generalized cross entropy loss for training deep neural networks with
  noisy labels.
\newblock In \emph{32nd Conference on Neural Information Processing Systems
  (NeurIPS)}, 2018.

\end{thebibliography}


\clearpage
\appendix

\thispagestyle{empty}

\setcounter{lemma}{0}
\setcounter{assumption}{0}
\setcounter{corollary}{0}
\setcounter{theorem}{0}

\twocolumn[ \makesupplementtitle ]

\setcounter{lemma}{0}
\setcounter{assumption}{0}
\setcounter{corollary}{0}
\setcounter{theorem}{0}

\section{Theoretical Results (Supplement)}
\label{appendix:theory}

This appendix is an expanded explanation of our theoretical results from Section \ref{sec:theory} in the main paper.
We begin with a short background explaining the standard properties of stochastic gradient descent that our proofs rely on,
then we restate our theoretical results from before along with detailed proofs.

\ignore{
\subsection{Summary}

It is well known that when using the standard cross entropy loss,
SGD converges at a rate of $O(\sqrt{k/n})$.
The tree loss's parameter sharing scheme asymptotically improves this convergence rate.

We assume that the class labels have a metric structure and let $c$ denote the doubling dimension of this metric.
This $c$ is a measure of the difficulty of our problem,
with larger $c$ implying more complex metrics and a more difficult learning problem.
This is captured in our convergence rates,
which show show that the tree loss converges at a rate of $O(\sqrt{\log k/n})$ for $c\le1$ or $O(\sqrt{k^{1-1/c}/n})$ for $c>1$.

But what happens if the tree is poorly constructed?
Remarkably, nothing bad.
We also show that for arbitrarily bad label tree structures,
the $U$ and $V$ losses maintain a $O(\sqrt{k/n})$ convergence rate.
This suggests that there is essentially no downside to using the tree loss parameterizations.

Our analysis relies on standard properties of stochastic gradient descent.
In this section, we first review these results,
then we formally state and prove our main result.
}

\subsection{SGD Background}

SGD is a standard algorithm for optimizing convex losses,
and it has many variations.
Our analysis uses the results from Chapter 14 of \citet{shalev2014understanding} which are based off of a version of SGD that uses parameter averaging.\footnote{
Other variants of SGD have similar results,
and we refer the reader to \citet{shalev2014understanding} for a discussion of these variants.}
In this section we review these results.
We follow the notation of \citet{shalev2014understanding} as closely as possible,
but we occasionally change the names of variables to avoid conflicts with variable names used elsewhere in this paper.\footnote{
    Specifically, we replace all Latin variables with Greek alternatives.
    We use $\theta$ instead of $\w$, $\nu$ instead of $\vv$, and $\beta$ instead of $B$.
    The Greek letters $\rho$ and $\eta$ have the same meaning in both documents.
}

SGD is an iterative procedure for optimizing a function $f : \Theta \to \mathbb R$ given only noisy samples of the function's gradient.
We let $\theta^{(t)}$ denote the parameter value at iteration $t$,
and $\eta : \mathbb R$ be the step size.
Let $\nu_t$ be a noisy gradient of $f$ satisfying
\begin{equation}
    \E[\nu_t | \theta^{(t)}] = \nabla f(\theta^{(t)}).
\end{equation}
Then the parameter update for step $t$ is defined to be
\begin{equation}
    \theta^{(t+1)} = \theta^{(t)} - \eta \nabla \nu_t
    .
\end{equation}
When using parameter averaging, the final model parameters are defined to be
\begin{equation}
    \bar\theta = \frac1T \sum_{i=1}^T \theta^{(t)}
    .
\end{equation}
Parameter averaging is not often used in practice because it has a large memory overhead,
but it is widely used in theoretical analysis of SGD.
Theorems using parameter averaging are typically much easier to state and prove,
and equivalent results without parameter averaging typically require an additional $\log T$ factor in their convergence rates \citep{rakhlin2011making}.

The main result we rely on is a bound on the convergence rate of SGD for convex Lipschitz functions.
We reproduce Theorem 14.8 of \citet{shalev2014understanding} in the theorem below.

\begin{theorem}
    \label{thm:sgd}
    Let $\beta,\rho>0$.
    Let $f$ be a convex, $\rho$-Lipschitz function and let $\star \theta = \argmin_{\theta : \ltwo{\theta}\le \beta} f(\theta)$.
    Assume that SGD is run for $T$ iterations with $\eta = \sqrt{\frac{\beta^2}{\rho^2 T}}$.
    Then,
    \begin{equation}
        \E f(\bar \theta) - f(\star \theta) \le \frac{\beta\rho}{\sqrt{T}}
        .
    \end{equation}
\end{theorem}

\subsection{SGD and Statistical Learning}

We now introduce more statistical learning theory notation,
and apply Theorem \ref{thm:sgd} above to bound the rate of learning.

Define the true loss of our learning problem as
\begin{equation}
    L_D(A) = \E_{(\x,y)\sim D} \ell(A; (\x, y))
\end{equation}
where $A \in \{W, U, V\}$ is a parameterization of the cross entropy loss.
The optimal parameter vector is defined to be the minimum of the true loss:
\begin{equation}
    \star A = \argmin_{A} L_D(A).
\end{equation}
In order to apply Theorem \ref{thm:sgd} to a learning problem,
we make the following assumption about the size of our data points.
\begin{assumption}
    \label{ass:lip}
    For each feature vector $\x$ in the data set, 
        $\ltwo{\x} \le \rho$
        .
\end{assumption}
Under this assumption, the cross entropy loss $\ell$ is $\rho$-Lipschitz for all parameterizations,
and the true loss $L_D$ is therefore also $\rho$-Lipschitz.

We use SGD to optimize $L_D$.
We let $f=L_D$, and $\theta = A$,
so $\nu_t$ is the gradient of $\ell$ on the $t$th data point.
Applying Theorem \ref{thm:sgd} gives the following convergence bound.
\begin{lemma}
    \label{ref:cor:A}
    Under Assumption \ref{ass:lip},
    we have that for any parameterization $A$ of the cross entropy loss,
    SGD converges at the rate
    \begin{equation}
        \E L_D(\bar A) - L_D(\star A) \le \frac{\lF{\star A}\rho}{\sqrt n}.
        \label{eq:Aconv}
    \end{equation}
\end{lemma}

We emphasize that this convergence rate above is both computational and statistical.
The expression on the left hand side of \eqref{eq:Aconv} is equivalent to the expected generalization error of the parameter $\bar A$,
and so we use the terms ``convergence rate'' and ``generalization error'' essentially interchangeably.

The important intuition of Lemma \ref{ref:cor:A} is that the generalization error of the cross entropy loss depends on the Frobenious norm of the optimal parameter matrix.
We can therefore compare the convergence of the standard parameterization with the $U$/$V$-tree parameterizations by comparing these norms.

Our next task is to recover the standard convergence rate of multi-class classification, and so we will now bound $\lF{\star W}$.
We make the following assumption.
\begin{assumption}
    \label{ass:B}
    For each class $i$, the optimal parameter vector $\star\w_i$ satisfies
        $\ltwo{\star\w_i} \le B$
        .
\end{assumption}
%
It follows that
\begin{equation}
    \lF{\star W}^2 = \sum_{i=1}^k \ltwo{\star\w_i} \le kB^2.
    \label{eq:starW}
\end{equation}
Substituting Eq \eqref{eq:starW} into \eqref{eq:Aconv} gives the following bound.
\begin{corollary}
\label{theorem:xentropy}
    Under assumptions \ref{ass:lip} and \ref{ass:B},
    then the generalization error of the standard cross entropy parameterization when trained with SGD is
\begin{equation}
    \E L_D(\bar W) - L_D(W^*)
    \le \frac {\sqrt kB\rho}{\sqrt n}
    .
\end{equation}
\end{corollary}

\ignore{
It is important to note that this result above is an upper bound and not a lower bound.
This $O(\sqrt{k/n})$ convergence rate matches the convergence rate implied by the Nataranj dimension \citep{}.
We are not aware of any lower bounds for SGD in the Lipschitz setting.
\citet{nguyen2018tight,jentzen2020lower} study the strongly convex case and find that the upper bounds match the lower bounds.
}

\subsection{Bounding the Tree Loss}

We now bound the Frobenius norm of $\star U$ in order to bound the generalization of the $U$-tree loss.
By construction, $\lF{\star V} \le \lF{\star U}$,
so this will also bound the convergence rates of the $V$-tree loss.

The analysis is divided into two parts.
First, we make no assumptions that the $U$-tree structure is good,
and we show that even in the worst case $\lF{\star U} = O(\lF{\star W})$.
This implies that using the tree loss cannot significantly hurt our performance.
\begin{lemma}
    \label{lemma:starU}
    Under assumption \ref{ass:B},
    the following bound holds for all tree structures:
    \begin{equation}
        \lF{\star V} \le \lF{\star U} \le 2\sqrt{k}B.
    \end{equation}
\end{lemma}
\begin{proof}
    We have
    \begin{align}
        \lF{\star U}^2 
        &= \sum_{i=1}^k \ltwo{\star\uu_i}^2 \\
        &= \sum_{i=1}^k \ltwo{\star\w_i - \star\w_{\parent{i}}}^2 \\
        &\le \sum_{i=1}^k \left(\ltwo{\star\w_i} + \ltwo{\star\w_{\parent{i}}} \right)^2 \\
        &\le \sum_{i=1}^k (2B)^2 \\
        &\le k (2B)^2
        .
    \end{align}
\end{proof}
Now we consider the more important case of when we have a useful $U$-tree structure that meaningfully captures the similarity between classes.
This idea is captured in the following assumption.
\begin{assumption}
    \label{ass:metric}
    Let $\lambda \ge 1$, and $d$ be a distance metric over the labels such that for all labels $i$ and $j$,
\begin{equation}
    \frac 1 \lambda d(i,j)
    \le \ltwo{\star \w_i - \star \w_j}
    \le \lambda d(i, j).
\end{equation}
We denote by $c$ the doubling dimension of the metric $d$,
and we assume that the $U$-tree structure is built using a cover tree \citep{beygelzimer2006cover}.
\end{assumption}

The Cover Tree was originally designed for speeding up nearest neighbor queries in arbitrary metric spaces.
The definition is rather technical, so we do not restate it here.
Instead, we mention only two important properties of the cover tree.
First, it can be constructed in time $O(k)$,
so building the cover tree is a cheap operation that has no meaningful impact on training performance.
It also only has to be done in the initialization step of the learning process, and not on every iteration.
Second, the cover tree construction has a hyperparameter which we call \texttt{base}.
This hyperparameter controls the fanout and depth of the tree because at any node $i$ at depth $\depth{i}$ in the tree,
the cover tree maintains the invariant that $d(i, \parent{i}) \le \texttt{base}^{-\depth{i}}$.
Increasing the $\texttt{base}$ hyperparameter results in shorter, fatter trees,
and decreasing it results in taller, narrower trees.
Our analysis follows the convention of setting $\texttt{base}=2$,
but we show in the experiments below that good performance is achieved for a wide range of base values.

The following Lemma uses Assumption \ref{ass:metric} and properties of the cover tree to bound the norm of $\lF{\star U}$.
It is our main technical result and the proof is deferred to the appendix.

\begin{lemma}
    \label{lemma:main}
    Under Assumptions \ref{ass:B} and \ref{ass:metric},
    when $c\le1$, we have that
    \begin{equation}
        \lF{\star U} \le \tfrac{1}{\sqrt2}\lambda B \sqrt{\log_2 k},
        \label{eq:c<=1}
    \end{equation}
    and when $c>1$, we have that
    \begin{equation}
        \lF{\star U} \le \sqrt{5}\lambda B \sqrt{k^{(1-1/c)}}.
        \label{eq:c>1}
    \end{equation}
\end{lemma}

\begin{proof}
    First, we review some simple properties of the cover tree structure that will be needed in our proof.
    Let $q$ denote the maximum number of children of any node in the cover tree.
    A standard property of cover trees is that $q \le (4^c)$ \citep[Lemma 25]{izbickithesis}.
    Let $Q_i$ denote the number of nodes at level $i$ in the cover tree.
    Then we have that $|Q_i| \le q^i.$
    Finally, let $r$ be the height of the cover tree.
    Without loss of generality, we analyze the bound when the cover tree is perfectly balanced and so $r = \log_q k = \log_2 k / (2c)$.

We have that
\begin{align}
    \lF{\star U}^2
    &= \sum_{i=1}^k \ltwo{\star\uu_i}^2 \\
    &= \sum_{i=0}^r \sum_{j\in Q_i} \ltwo{\star\uu_j}^2 \\
    &= \sum_{i=0}^r \sum_{j\in Q_i} \ltwo{\star\w_j - \star\w_{\parent{j}}}^2 \\
    &\le \sum_{i=0}^r \sum_{j\in Q_i} \lambda^2 d(j, \parent{j})^2 \\
    &\le \sum_{i=0}^r \sum_{j\in Q_i} \lambda^2 (B/2^i)^2 \\
    &\le \sum_{i=0}^r q^i \lambda^2 (B/2^i)^2 \\
    &\le \sum_{i=0}^r \lambda^2 B^2 (q/4)^i 
    \label{eq:precases}
\end{align}
We proceed by considering the case of $c\le1$ and $c>1$ separately.

When $c\le1$, $q\le4$.
Substituting into Eq \eqref{eq:precases}, we have
\begin{align}
    \lF{\star U}^2
    &\le \sum_{i=0}^r \lambda^2 B^2 \\
    &= (r+1) \lambda^2 B^2 \\
    &= \frac{\log_2k}{2c} \lambda^2 B^2 \\
    &\le \frac{\log_2k}{2} \lambda^2 B^2
    .
\end{align}
This matches Eq \eqref{eq:c<=1} in the lemma's statement.

When $c>$, then $q>4$.
We first show the following equality.
\begin{align}
\left(\frac q 4\right)^r
&=
\left(\frac q 4\right)^{\log_q k}\\
&=
\frac{k}{4^{\log_q k}}\\
&=
\frac{k}{4^{{\log_4 k}/{\log_4 q}}}\\
&=
\frac{k}{k^{1/\log_4 q}}\\
&=
k^{\left(1 - {1}/{\log_4 q}\right)} \\
&=
k^{\left(1 - 1/c\right)}
.
\label{eq:lFU:frac}
\end{align}
Substituting into Eq \eqref{eq:precases}, we have
\begin{align}
    \lF{\star U}^2
    &= \lambda^2B^2 \left(\frac{(q/4)^{r+1}-1}{q/4-1}\right) \\
    \label{eq:lFU:4}
    &= \lambda^2B^2 \left(\frac{(q/4)k^{(1-1/c)}-1}{q/4-1}\right) \\
    &\le \lambda^2B^2 \left(\frac{(q/4)k^{(1-1/c)}}{q/4-1}\right) \\
    &\le 5 \lambda^2B^2 k^{(1-1/c)}
    \label{eq:lFU:5}
\end{align}
    Inequality \eqref{eq:lFU:5} follows because $\tfrac{q/4}{q/4-1} \le 5$ for integer values of $q>4$.
\end{proof}
We note that embedding techniques can be used to reduce the intrinsic dimension of the metric ($c$) at the expense of increasing the metric's distortion,
but we make no attempt to optimize this tradeoff in this paper.

We now state our main result.
It is an immediate consequence of the fact that $\lF{\star V} \le \lF{\star U}$ and Lemmas \ref{ref:cor:A} and \ref{lemma:main}.
\begin{corollary}
    \label{cor:main}
    Under assumptions \ref{ass:lip}, \ref{ass:B}, and \ref{ass:metric},
    when $c\le1$, the generalization error of the tree loss is bounded by
\begin{equation}
    \E L_D(\bar V) - L_D(V^*)
    \le \frac {\lambda B\rho \sqrt{\log_2 k}}{\sqrt 2n}
    .
\end{equation}
    When $c>1$, the generalization error of the tree loss is bounded by
\begin{equation}
    \E L_D(\bar V) - L_D(V^*)
    \le \frac {\lambda B\rho \sqrt{5 k^{(1-1/c)}}}{\sqrt n}
    .
\end{equation}
\end{corollary}
These convergence rates are asymptotically better than the convergence rates for the standard parameterization of the cross entropy loss.

\end{document}